\documentclass{article}
\usepackage[a4paper, total={6in, 8in}]{geometry}
\usepackage[utf8]{inputenc}
\usepackage{graphicx}
\usepackage{caption}
\usepackage{subcaption}
\usepackage{indentfirst}
\usepackage{amsmath,amssymb,amsthm}
\usepackage{listings}
\usepackage{graphics}

\usepackage{amsmath,amssymb,amsthm}
\usepackage[ruled,vlined]{algorithm2e}
\usepackage{caption}
\usepackage{subcaption}

\usepackage{xurl}
\usepackage[colorlinks,allcolors=blue]{hyperref}

\usepackage{color}
\usepackage[dvipsnames]{xcolor}

\title{Benford's law: what does it say on adversarial images?}
\author{João G. Zago\footnote{joao.zago@posgrad.ufsc.br}, Eric A. Antonelo\footnote{eric.antonelo@ufsc.br}, Fabio L. Baldissera, Rodrigo T. Saad}

\date{March, 2023}

\begin{document}

\maketitle

\begin{abstract}
Convolutional neural networks (CNNs) are fragile to small perturbations in the input images. 
These networks are thus prone to malicious attacks that perturb the inputs to force a misclassification. 
Such slightly manipulated images aimed at deceiving the classifier are known as adversarial images. 
In this work, we investigate statistical differences between natural images and adversarial ones.
More precisely, we show that employing a proper image transformation for a class of adversarial attacks, the distribution of the leading digit of the pixels in adversarial images deviates from Benford's law.
The stronger the attack, the more distant the resulting distribution is from Benford's law.
Our analysis provides a detailed investigation of this new approach that can serve as a basis for alternative adversarial example detection methods that do not need to modify the original CNN classifier neither work on the high-dimensional pixel space for features to defend against attacks.
\end{abstract}

\section{Introduction}
\label{intro}

Convolutional neural networks (CNN) are highly successful in image classification tasks \cite{szegedy2016inception}. 
However, they are not robust to small perturbations in their inputs \cite{szegedy2013intriguing, goodfellow2014explaining, wang2019sample}, i.e., slight changes in the pixel values of an input image might result in a different classification. 
Malicious attacks can explore this fragility of neural networks, giving rise to the so-called adversarial images \cite{goodfellow2014explaining}. 
The identification difficulty of manipulated images raises concerns for the application of neural networks in domains where safety is of primary interest \cite{huang2017safety,katz2017reluplex}. 

There are several different approaches proposed in the literature to address this problem. 
Grouped into two major categories:
a) network-centered, whose aim is to decrease the neural network vulnerability to adversarial images \cite{szegedy2013intriguing, goodfellow2014explaining, madry2017towards, tramer2017ensemble, kurakin2016adversarial,papernot2016distillation, LI2021103037};
b) input-centered, where the goal is to detect adversarial images \cite{katz2017reluplex, song2017pixeldefend, grosse2017statistical, metzen2017detecting, lu2017safetynet, yang2020ml, tian2021DetectingAE, chou2020DetectionLocUniversal, MAZUMDAR2022103417}.

This paper is related to approach (b) because we propose the use of Benford's Law (BL), also known as the First Digit Law (FDL), to expose adversarial images. After all, the findings we present here could be used to detect adversarial images.
BL states the behavior of the first digit distribution from natural datasets. 
According to this law, the leading digit distribution of real-world data follows a logarithmic function, described later in detail.
The successful use of BL in other domains, such as in image forensics, to detect frauds \cite{todter2009benford, deckert2011benford}, and image compression \cite{pevny2008detection, milani2016phylogenetic}, for instance, inspired the idea of applying BL in the context of adversarial image recognition.

Here we present that adversarial images devised by state-of-the-art attack algorithms display a leading digit distribution of the pixel values that deviate from those of natural images. 
While natural images seem to adhere to the FDL, the same is not valid for their corresponding perturbed images.

To the best of our knowledge, this is the first application of BL for this purpose. Our main contribution in this paper is to provide a solid empirical analysis that leads to the following claims: 
\begin{itemize}
    \item adversarial images, different from natural ones, tend to deviate significantly from BL;
    \item this deviation is higher for attack algorithms based on infinite-norm perturbations;
    \item deviations from Benford's Law increase with the magnitude of the generated perturbation;
    \item in some cases, adversarial attacks can be anticipated even before the perturbed image becomes adversarial, that is, a deviation from BL takes place during the formation of an attack; and
    \item another fundamental characteristic of this new approach is that it produces a computationally cheap low-dimensional input feature that could be used for adversarial image detection.
\end{itemize}

We organized this work as follows. Section \ref{sec:methods} presents the proposed approach to compute the deviation of adversarial and natural images concerning the distribution given by BL.
In Section \ref{sec:experiments}, we describe the experimental setup used to generate the data that supports our claims.
Major results are presented in Section \ref{sec:results}, while conclusions and future perspectives are given in Section \ref{sec:conclusion}.

\section{Methods} \label{sec:methods}

\begin{figure*}[t]
    \centering
    \includegraphics[scale=0.42]{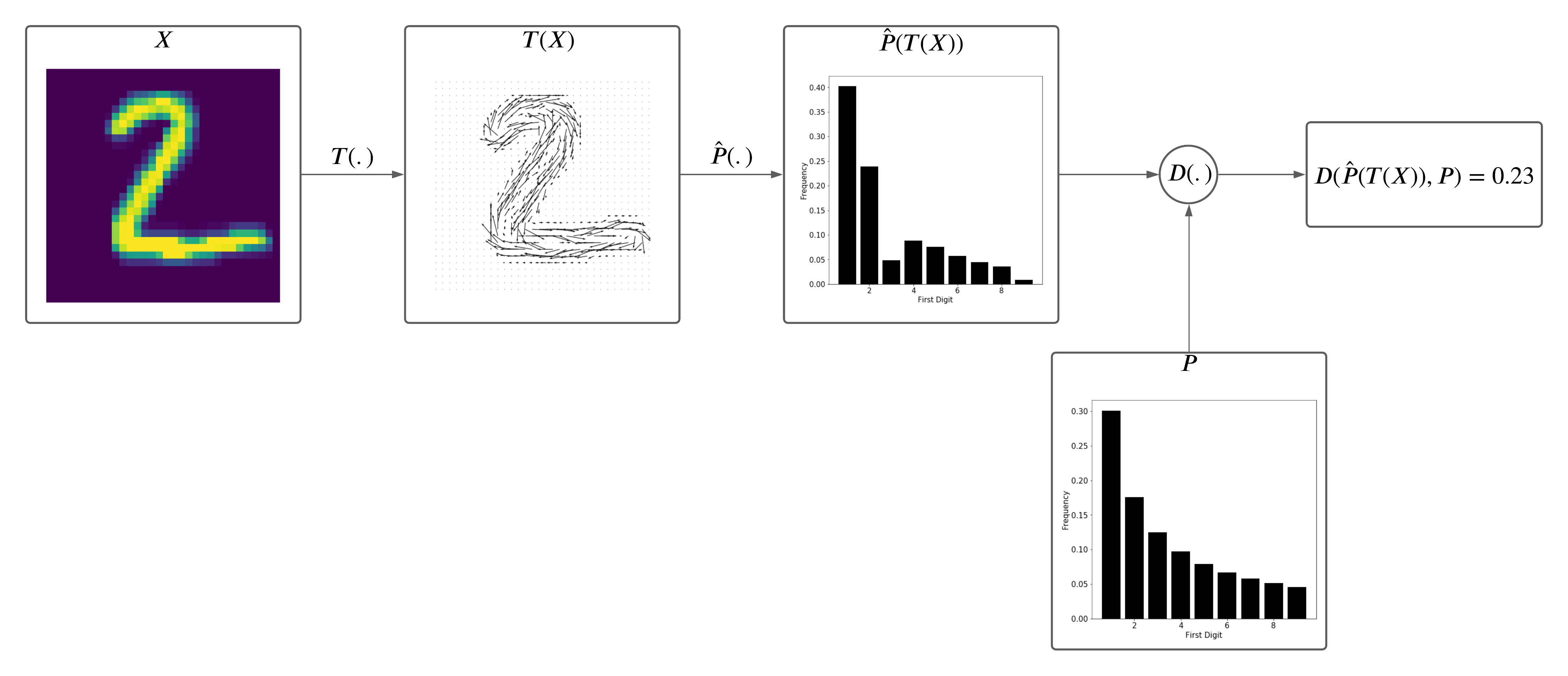}
    \caption{Overview of the proposed approach, composed of three steps: a) transformation of the image $x$, represented by the mapping $T(.)$; b) a statistical analysis of $T(x)$, denoted by $\hat{P}$ and c) a comparison of $\hat{P}$ with a distribution of reference $P$.}
    \label{fig:method_representation}
\end{figure*}

In this paper, we may refer to an input image $x$ both as a vector ($x\in\mathbb{R}^{n}$) or as a 2-D matrix ($x\in\mathbb{R}^{n \times m}$), to simplify notation. Given any image $x$, we summarize our approach as follows:
a) compute the gradient of $x$, here denoted $T(x)$; then
b) get the frequency of the first digits of $T(x)$; 
c) compare, through the Kolmogorov-Smirnov test, the distribution got in (b) with the one given by Benford’s law. 
We present the procedure encompassed by steps (a)-(c) in Figure \ref{fig:method_representation}. We now detail each of these steps in the sections that follow.

\subsection{Benford's Law}

Benford's Law or the First Digit Law states that, across different domains, the distribution of the leading digits of numerical data follows a similar pattern, namely, the one given by Equation \ref{eq:benfords_eq}:
\begin{equation}
    P(d) = \log_{10} \left(1 + \frac{1}{d}\right)
    \label{eq:benfords_eq}
\end{equation}

\begin{figure}[ht]
    \centering
    \includegraphics[scale=0.40]{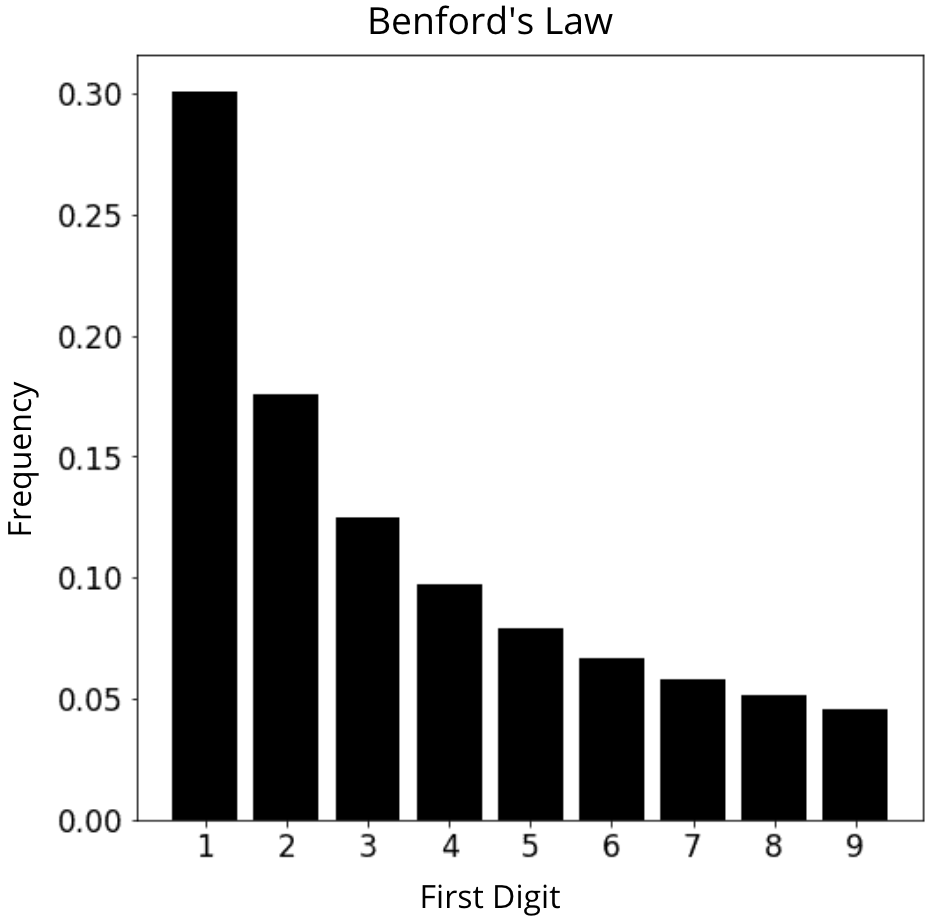}
    \caption{First digit distribution as in Benford's Law}
    \label{fig:benfords_fig}
\end{figure}

Figure \ref{fig:benfords_fig} portrays the First Digit distribution, as proposed in BL.

Pixel-based images rarely follow BL \cite{jolion2001images}. However, for certain transformations, the transformed images do. Two examples of such transformations are: the gradient magnitude of the input image \cite{jolion2001images} and the Discrete Cosine Transformation (DCT) \cite{perez2007benford}.
We employed the former, presented next, to map images such that the natural ones behave as in BL.

\subsection{Image Transformation: Gradient Magnitude}

Given an input image $x \in \mathbb{R}^{n \times m}$, we compute its gradient $G(x)$ according to Equation \ref{eq:gradient} below:
\begin{equation}
    G(x)_{i,j} = \sqrt{G_{e_1}(x)_{i,j}^2 + G_{e_2}(x)_{i,j}^2}, 
    \label{eq:gradient}
\end{equation}
where $G(x)$ is the gradient magnitude; $i$ and $j$ are indices indicating each pixel value; $G_{e_1}(x)$ and $G_{e_2}(x)$ are the horizontal and vertical components of the gradient approximation, given by:
\begin{equation}
    G_{e_1}(x) = x * K_{e_1}  \text{, }
    G_{e_2}(x) = x * K_{e_2}
    \label{eq:sobel_grad}
\end{equation}
which are computed using the following Sobel filters $K_{e_1}$ and $K_{e_2}$ for the convolution operation with the input image:

\begin{equation}
  K_{e_1}=
  \begin{bmatrix}
    -1 & 0 & 1 \\
    -2 & 0 & 2 \\
    -1 & 0 & 1
  \end{bmatrix} \text{, }
  K_{e_2}=
  \begin{bmatrix}
    -1 & -2 & -1 \\
    0 & 0 & 0 \\
    1 & 2 & 1
  \end{bmatrix}
  \notag
\end{equation}

The discrete convolution operation, employed in Equation \ref{eq:sobel_grad} for the approximation of the gradient in both directions, is given by:
\begin{equation}
    (x * K)_{i,j} = \sum_{o=1}^{O} \sum_{p=1}^{P} x_{i - o, j - p}  K_{o,p}
    \label{eq:discrete_convolution}
\end{equation} 
where $K$ represents, generically, both of the Sobel filters; $O$ and $P$ are the total number of rows and columns, respectively, relative to $K$.

\subsection{First Digit Distribution (FDD)}

Given the transformed image $T(x)$, we aim to calculate its associated First Digit Distribution (FDD). Firstly, we get the leading digit of each of its pixel values.
The first digits of $T(x)$ are denoted as $F(T(x))$
(e.g., $F$ will output 1 given the pixel value 176 and 5 given 54).
Then, the frequency for each digit $d \in \{1,\ldots,9\}$ is computed, forming a distribution $\hat{P}(d)$, satisfying $\sum_{d=1}^9 \hat{P}(d) = 1$.
Note that $T(x)$ corresponds to $G(x)$ in this work.

\subsection{Deviation between distributions: the Kolmogorov-Smirnov (KS) test}

Given two distributions, one may use the Kolmogorov-Smirnov (KS) test to compute how much these distributions diverge. 
This test evaluates the distance between two empirical distributions or between the theoretical and an empirical distribution.

To assess the difference between the empirical distribution $\hat{P}$ and the theoretical distribution $P$, given by BL, we employed this test by calculating the KS statistic between these distributions, denoted as $D^{KS}(\hat{P},P)$, using the formula below:
\begin{equation}
    D^{KS}(p,q) = \sup|Acc(p) - Acc(q)|,
    \label{eq:ks-test}
\end{equation}
where $Acc$ returns the accumulated distribution of its given input distribution; and $p$ and $q$ correspond to the distributions.

\section{Experimental Setup} \label{sec:experiments}

Here, we present the setup employed to analyze the deviation of adversarial images from Benford's distribution.
We briefly present the selected image datasets as well as the employed CNN architecture and training parameters.
We also describe the adversarial attacks employed in our experiments.

\subsection{Datasets}
 
We considered three different image datasets in this work: 

\begin{itemize}
    \item MNIST: data set containing grayscale images for digits from 0 to 9 \cite{mnisthandwrittendigit2010};
    \item CIFAR10: data set composed of 10 different classes of RGB images  \cite{krizhevsky2009learning};
    \item ImageNet: data set composed of 1000 different classes of RGB images \cite{imagenet_cvpr09}.
\end{itemize}

\subsection{CNNs under attack}

We employed a different CNN model architecture regarding each data set.
The CNNs employed to classify images from the MNIST (Table \ref{tab:neural_network_architecture_mnist}) and CIFAR10 (VGG16 \cite{simonyan2014very}) datasets were trained from scratch. We employed a benchmark network (already trained) for the  ImageNet data (VGG19 \cite{simonyan2014very}). 

For the MNIST data set, we employed the ADAM \cite{kingma2014adam} optimizer during the training procedure with a learning rate of $1\mathrm{e}{-3}$, $\beta_1 = 0.9$, $\beta_2 = 0.999$ and $\epsilon = 1\mathrm{e}{-08}$.
The train, test, and validation set comprised: $40,000$, $10,000$ and $10,000$ images, respectively.
The classifier was trained for about $50$ epochs, reaching a training accuracy of about $99.29\%$ and for the test set $97.03\%$.

\begin{table}[htbp]
    \centering
    \caption{CNN architecture under attack for MNIST dataset}
    \begin{tabular}{|c|c|c|}   
        \hline
        Layer & Type & Dimensions \\ \hline
        0 & Input & (32x32x3) \\ \hline
        1 & Conv(3x3)-64 & (32x32x64) \\ \hline
        2 & Conv(3x3)-64 & (32x32x64) \\ \hline
        3 & Batch Normalization & (32x32x64) \\ \hline
        4 & Max Pooling(2x2) & (16x16x64) \\ \hline
        5 & Conv(3x3)-128 & (16x16x128) \\ \hline
        6 & Conv(3x3)-128 & (16x16x128) \\ \hline
        7 & Batch Normalization & (16x16x128) \\ \hline
        8 & Max Pooling(2x2) & (8x8x128) \\ \hline
        9 & Conv(3x3)-256 & (8x8x256) \\ \hline
        10 & Conv(3x3)-256 & (8x8x256) \\ \hline
        11 & Batch Normalization & (8x8x256) \\ \hline
        12 & Max Pooling(2x2) & (4x4x256) \\\hline
        13 & Fully Connected & (2048) \\ \hline
        14 & Batch Normalization & (2048) \\ \hline
        15 & Fully Connected & (2048) \\ \hline
        16 & Fully Connected & (10) \\\hline
    \end{tabular}
    \label{tab:neural_network_architecture_mnist}
\end{table}

For the CIFAR10 classifier, the optimization algorithm employed in the training procedure was the Gradient Descent, with a learning rate of $1\mathrm{e}{-3}$ and momentum of $0.9$.
The train, validation and test sets were composed of: $40,000$, $10,000$ and $10,000$, respectively.
The training procedure took about $50$ epochs, reaching a training accuracy of about $99.74\%$ and for the test set $80.11\%$.

\subsection{Adversarial Attacks}

We employed three different adversarial attack algorithms, the Fast Gradient Sign Method (FGSM) \cite{goodfellow2014explaining}, the Projected Gradient Descent (PGD) \cite{madry2017towards} and the Carlini and Wagner (C\&W) \cite{carlini2017towards}. Each algorithm generated the adversarial examples associated with each CNN classifiers presented in the previous section. All of them are white-box attacks \cite{yuan2019adversarial}.

The FGSM attack \cite{goodfellow2014explaining} was designed to generate adversarial examples in a one-step process. This attack method uses internal information from the classifier and has access to the training set to craft perturbed images.
The algorithm computes the perturbation by applying the gradient of the loss function $J(\theta, x,y)$, which is usually the same employed for training a neural net classifier, where $x$ is a given input image and $y$ the target output; $\theta$ represents the weights of a neural net.
But, here, the gradient of the loss function is taken regarding the input image $x$.
Then, $x$ is perturbed as follows: 

\begin{equation}
    x^* = x + \epsilon \text{ } sign(\nabla_{x} J(\theta, x, y)), 
    \label{eq:fgsm}
\end{equation}
where $\epsilon$ is the magnitude of the perturbation and $x^*$ is the new image which was perturbed using only the sign of the gradient
in a direction to maximize the loss function.

The difference between FGSM and PGD is that the latter is an iterative method, consisting of a series of updates to the input image under attack:
\begin{equation}
    x_{i+1} = x_{i} + \epsilon \text{ } sign(\nabla_{x_{i}} J(\theta, x_{i}, y)),
    \label{eq:pgd}
\end{equation}
where $i$ stands for the iteration index.
The PGD aims to generate adversarial examples with a small perturbation, though it is slower due to the iterative process. Similarly to PGD, the C\&W attack also uses an iterative approach to perform the attack, but it employs a different objective function \cite{carlini2017towards}.

\subsection{Experimental Description} \label{sec:experimental_setup}

Before measuring the deviation between the FDD of the original images and the FDD of the adversarial images, we employed the following steps:

\begin{enumerate}
    \item Select 1000 random images from the test set for each of the datasets: MNIST, CIFAR10, and ImageNet;
    \item Attack each of these selected images with one of the selected adversarial attack algorithms (FGSM, PGD, or C\&W);
    \item Transform both the original and the adversarial images with the gradient magnitude method using Equation (\ref{eq:gradient});
    \item Compute the resulting first digit distribution (FDD) $\hat{P}$ for each transformed image;
\end{enumerate}

With the FDD of all images already computed, we can use the KS statistic or KL divergence as follows:
\begin{enumerate}
    \item Apply the KS statistic for the FDD of the original unattacked images concerning the FDD from Benford's Law (Equation \ref{eq:ks-test}). Do the same for the adversarial images.
    \item Compute the divergence (using both, the Kullback-Leibler divergence and the Kolmogorov-Smirnov test) between the FDD of each clean image regarding the FDD from Benford's Law (Equation \ref{eq:kl-divergence}). Do the same for the adversarial images.
\end{enumerate}

The Kullback-Leibler (KL) divergence between distributions $p$ and $q$ is given as:
\begin{equation}
    D^{KL}(p,q) = \sum_{d = 1}^{9} p(d) \log(p(d) \div q(d))
    \label{eq:kl-divergence}
\end{equation}

\begin{table}[htbp]
    \centering
    \caption{CNN architecture for adversarial detection}
    \begin{tabular}{|c|c|c|}   
        \hline
        Layer & Type & Dimensions \\ \hline
        0 & Input & (224x224x3) \\ \hline
        1 & Conv(3x3)-32 & (224x224x32) \\ \hline
        2 & Max Pooling(2x2) & (112x112x32) \\ \hline
        3 & Batch Normalization & (112x112x32) \\ \hline
        4 & Conv(3x3)-32 & (112x112x32) \\ \hline
        5 & Max Pooling(2x2) & (56x56x32) \\ \hline
        6 & Batch Normalization & (56x56x32) \\ \hline
        7 & Conv(3x3)-32 & (56x56x32) \\ \hline
        8 & Max Pooling(2x2) & (28x28x32) \\ \hline
        9 & Batch Normalization & (28x28x32) \\ \hline
        10 & Conv(3x3)-32 & (28x28x32) \\ \hline
        11 & Max Pooling(2x2) & (14x14x32) \\ \hline
        12 & Batch Normalization & (14x14x32) \\ \hline
        13 & Conv(3x3)-32 & (14x14x32) \\ \hline
        14 & Max Pooling(2x2) & (7x7x32) \\ \hline
        15 & Batch Normalization & (7x7x32) \\ \hline
        16 & Fully Connected & (64) \\ \hline
        17 & Batch Normalization & (64) \\ \hline
        18 & Fully Connected & (64) \\ \hline
        19 & Batch Normalization & (64) \\ \hline
        20 & Fully Connected & (1) \\ \hline
    \end{tabular}
    \label{tab:neural_network_binary_architecture}
\end{table}

In addition to analyzing deviations between the FDD of images, we also developed a logistic regression classifier that uses the KS test output (the deviation between distributions of the input image and the theoretical one from BL) as a unidimensional feature as input. We compared the latter to the CNN (Table \ref{tab:neural_network_binary_architecture}) trained to detect an adversarial example based on the whole input image.
We created the dataset to train both classifiers as follows:

\begin{enumerate}
    \item Select 1,000 random images from the test set of the ImageNet dataset;
    \item Attack each of the selected images with both PGD approaches ($2-norm$ and $\infty-norm$);
    \item 
    Label 500 images as adversarial and 500 images as original (unattacked) for each PGD approach, with a resulting dataset of 1,000 images for each approach;
\end{enumerate}

The classifiers are trained for each PGD approach with the corresponding dataset and evaluated according to the f1-score, recall, precision, and accuracy metrics.

\section{Results} \label{sec:results}

\subsection{Adversarial images tend to significantly deviate from Benford's Law}

\begin{figure*}[!t]
    \begin{subfigure}{\textwidth}
        \centering
        \includegraphics[scale=0.5]{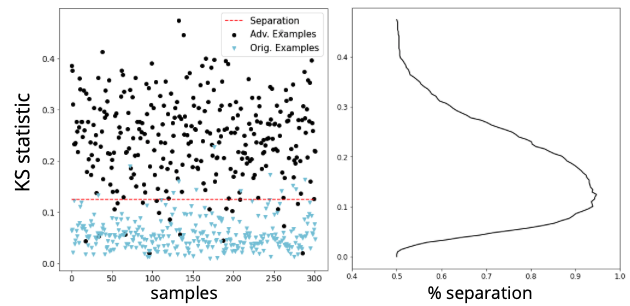}
        \caption{Represents the adversarial examples generated using the $||.||_\infty-norm$ PGD attack}
        \label{fig:kl_div_sep_norm_inf}
    \end{subfigure}
    
    \begin{subfigure}{\textwidth}
        \centering
        \includegraphics[scale=0.5]{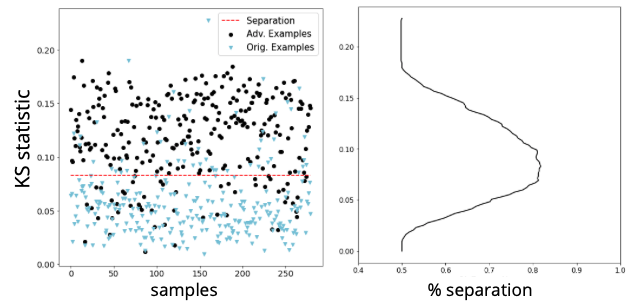}
        \caption{Results associated with the examples crafted with the $||.||_2-norm$ PGD attack}
        \label{fig:kl_div_sep_norm_2}
    \end{subfigure}
    \caption{
    Separation using the KS statistic for adversarial and clean examples. 
    Left: each dot represents one image, attacked (in black) or unattacked (in blue).
    The maximum separation is achieved by the red horizontal line.
    Right: separation percentage of the points from the left plot for different horizontal lines that split those points linearly. The maximum is attained by the horizontal red line.}
    \label{fig:kl_div_sep}
\end{figure*}

In Figure \ref{fig:kl_div_sep}, we can view the separation between adversarial images (black dots) and clean images (blue dots) achieved by just plotting the KS test value for all images. The PGD attack was applied using both the $||.||_\infty-norm$ (Figure \ref{fig:kl_div_sep}-a) and $||.||_2-norm$ approaches (Figure \ref{fig:kl_div_sep}-b).

Meaning that the  FDD of the adversarial images deviate more from the FDL than the FDD of the clean images, since the black dots are higher than the blue dots in the figure, for the chosen dataset and attack algorithm. We also note that it is possible to build a linear classifier using this one-dimensional feature, which is the output of the KS test.

In terms of separation percentage or \textit{classification} performance, 94.7\% of the images were correctly \textit{classified} for the $||.||_\infty-norm$ PGD attack, and 81.8\% of the images for the $||.||_2-norm$ PGD attack.

\subsection{Images generated by $||.||_\infty-norm$ attacks deviate more from the Benford's distribution than those created by $||.||_2-norm$ attacks}

In Figure 3, we can observe that the interval where the black dots (adversarial examples) are located is greater for the $||.||_\infty-norm$ attack in comparison with the $||.||_2-norm$ attack. For the former, the black dots reach up to a deviation of 0.4 from the BL's theoretical distribution, which is considerably higher than the deviation of approximately 0.17 for the latter attack. 
This result remains valid for other image datasets.

From the attacker's point of view, there is a trade-off in terms of robustness regarding adversarial examples between two different distance metrics ($||.||_\infty-norm$ and $||.||_2-norm$) \cite{Khoury2018OnTG}.
This effect can be seen in Figure 3, where the perturbed images from $||.||_2-norm$ PGD attacks are more difficult to be discriminated from the clean original images when compared to the $||.||_\infty-norm$ attack. That is, the \textit{classification} performance of a linear classifier using the KS test output as input feature is higher for the $||.||_\infty-norm$ attack than the $||.||_2-norm$ attack.

\subsection{The deviation from Benford's distribution increases with the magnitude of the attack's perturbation}

\begin{figure}[ht]
    \centering
    \includegraphics[scale=0.38]{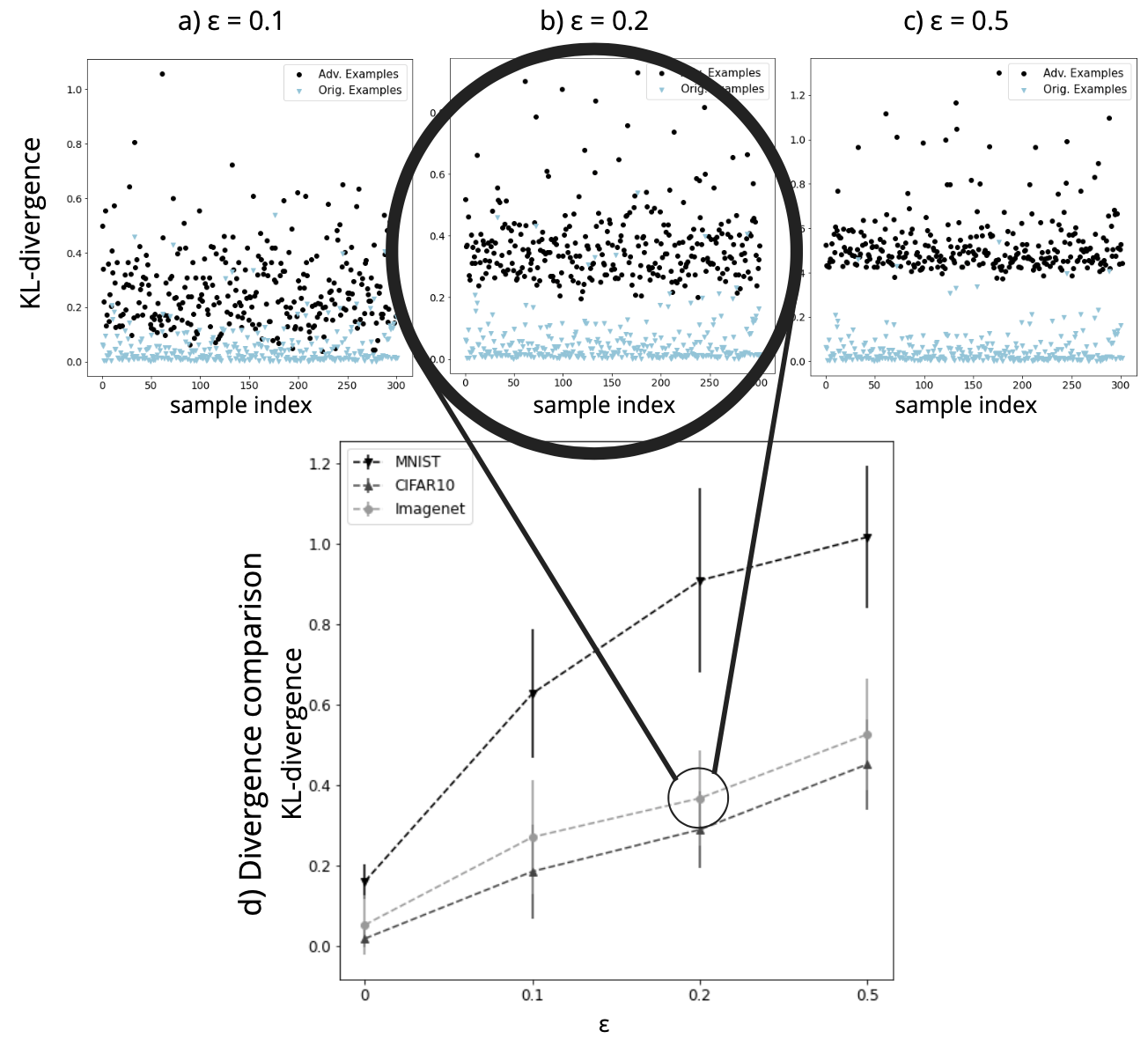}
    \caption{
    Separation between adversarial (black dots) and clean images (blue dots) increases with magnitude of attack's perturbation $\epsilon$.
    (a), (b), and (c) present the KL divergence between the FDD of the transformed images (samples from Imagenet dataset) and the FDL by Benford's Law.
    The adversarial examples are generated by the $||.||_\infty$-norm FGSM attack with $\epsilon$ equal to $0.1$, $0.2$ and $0.5$, respectively.
    (d) shows the mean and standard deviation for the KL-divergence obtained while varying $\epsilon$ for three different datasets.
    }
    \label{fig:kl_div_fgsm_inf}
\end{figure}

In Figure \ref{fig:kl_div_fgsm_inf}, we can see that the higher the attack's perturbation $\epsilon$ imposed on the images, the more distant the adversarial images become regarding the clean unattacked images. This behavior remains valid for all the datasets employed in the experiments.
Moreover, with higher $\epsilon$, it also becomes easier to detect adversarial images as the two classes of images become better separated in terms of the KL divergence.

\subsection{An attack can be anticipated as it is iteratively formed}

\begin{figure}[ht]
    \centering
    \includegraphics[scale=0.3]{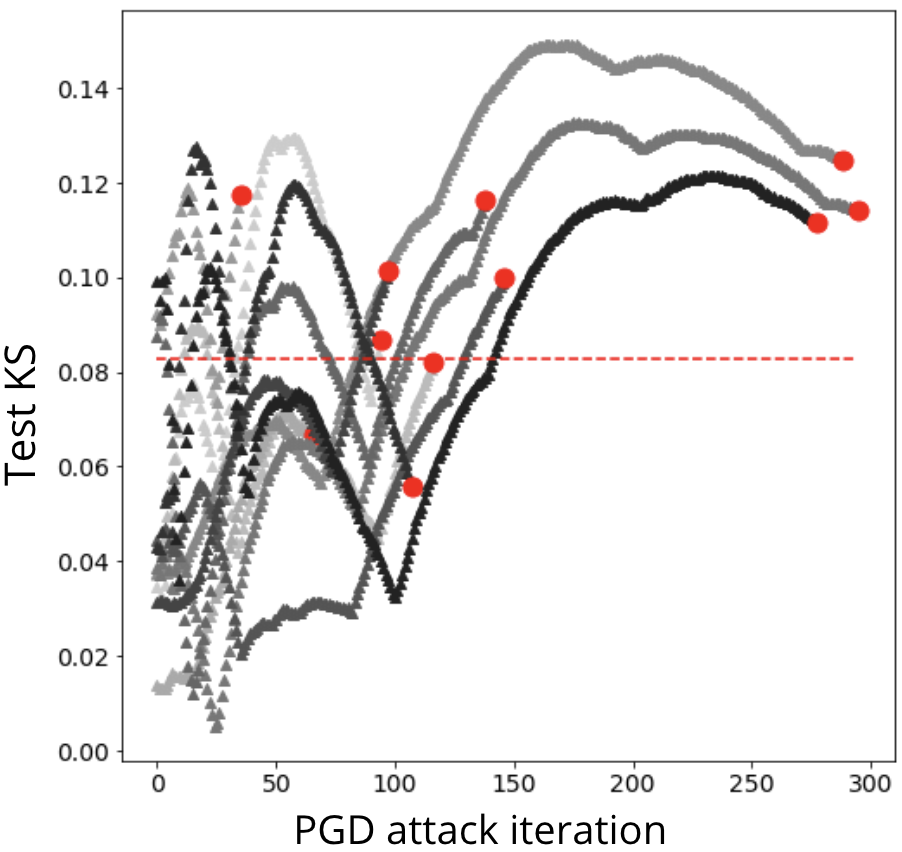}
    \caption{
    Output of the KS test (our proposal) as the $2-norm$ PGD attack is formed for eleven images from the Imagenet dataset. 
    Each trajectory represents one image that is being attacked, with the final adversarial image represented by a red dot in the end of the trajectory. The points (images) above the dashed horizontal red line can be flagged as \textit{under attack} by our method.
    }
    \label{fig:ks_x_iteration}
\end{figure}

Our method also helps monitor whether a neural network is under attack by computing the deviation given by the KS statistic for all images that the network would receive as input.
In Figure \ref{fig:ks_x_iteration}, we can observe eleven images under attack from the first iteration until the last iteration of the $2-norm$ PGD attack, when the image becomes adversarial (red dot). We can see that before this happens, the image under attack can already be preemptively flagged as \textit{under attack} or heading to become adversarial by our method since it already starts deviating from the theoretical distribution of Benford's law.
The dashed horizontal red line represents a decision boundary splitting both classes, adversarial and clean (found as in Figure 3).
We can see many grey points (images under attack) above that boundary, meaning that the deviation is enough to be flagged as \textit{under attack} before the attack end.
To the best of our knowledge, so far this is the first method that can perform in this way, i.e, be easily and inexpensively applied on an ongoing attack.

\subsection{KS test vs. KL divergence}

Here, we compare the KS test to the KL divergence as a procedure to compute deviations between distributions in our method.
We have chosen the PGD and C\&W attacks, as they are considered the strongest first-order attacks and have deceived most of the defense methods \cite{madry2017towards, carlini2017adversarialdetection}.
We noticed by the results presented in Table \ref{tab:ks_x_kl} that there is a significant increment in the separation percentage when applying the KS test under any of the considered approaches.

\begin{table}[bt]
    \caption{Maximum separation percentage comparison between the Kullback-Leibler divergence and the Kolmogorov-Smirnov statistics.}
    \centering
    \begin{tabular}{|l|l|l|}
        \hline
        Attack & KL-divergence & KS test \\
        \hline
        PGD $\infty$-norm & 90.23\% & 94.70\% \\
        \hline
        PGD $2$-norm & 66.96\% & 81.79\% \\
        \hline
        C\&W $2$-norm & 65.07\% & 83.28\% \\
        \hline
    \end{tabular}
    \label{tab:ks_x_kl}
\end{table}

This analysis suggests that the KS test is more sensitive to the deviations on the first digit distribution because it reaches a higher separation percentage using the same amount of information given for the KL divergence.

\subsection{Feature for Adversarial Detection (FAD)}

The output of our method, which is the deviation between distributions, can be used as a feature of any classifier. The goal of this classifier is to detect whether an image is adversarial or not.
We use simply a logistic regression classifier that has a uni-dimensional input coming from the KS test of our proposed method. The training set has 350 adversarial images and 350 clean unattacked images, whereas the test set has 300 as Section \ref{sec:experimental_setup} described.
There are two training sets: one for the $||.||_2$ PGD attack method and one for the $||.||_\infty$ PGD attack.
Tables \ref{tab:logistic_regression_metric} summarize the results.
We can see that the binary classifier trained to detect adversarial examples based on the proposed feature has an accuracy consistent with those values from the maximum separation percentage, presented in Table \ref{tab:ks_x_kl}. 
The test classification performance was 82\% for PGD $2-norm$ and 92\% for PGD $\infty-norm$.

\begin{table}[ht]
    \caption{Results for Logistic regression with proposed FAD using ImageNet examples.}
    \centering
    \begin{tabular}{|l|l|l|l|l|l|}
         \hline
         PGD & dataset & f1-score & precision & recall & acc. \\
         \hline
         $\infty$-norm & test & 0.91 & 0.94 & 0.88 & 0.92 \\
         & train & 0.93 & 0.98 & 0.88 & 0.94 \\
         \hline
         $2$-norm & test & 0.81 & 0.80 & 0.83 & 0.82 \\
         & train & 0.81 & 0.82 & 0.80 & 0.81 \\
         \hline
    \end{tabular}
    \label{tab:logistic_regression_metric}
\end{table}

Now, to compare the results with a common method in the literature, we train a CNN to do the same adversarial image detection, but now based on the whole raw image as input
as proposed in \cite{metzen2017detecting}.
The training and test sets remain the same as before, except for the input data, that comprises the whole image.
The performance achieved by our FAD (feature for adversarial detection) approach with logistic regression and by the CNN was very similar when considering the $\infty-norm$ attack. 
Although the CNN achieved a better result for the PGD $2-norm$, notice that our FAD approach uses a single variable as input, whereas the CNN  classifier uses the whole image (150,528 pixel inputs) and has 143,681 parameters. This difference makes the CNN much slower during either training or inference and more memory-demanding than our method.
Furthermore, our FAD is very general and already provides a linearly separable unidimensional input which indicates the deviating nature of an adversarial image, and more importantly, requires no training at all (since it is based solely on preprocessing/transformation steps of the input image).

\begin{table}[ht]
    \caption{Results for the CNN-based classifier with whole image as input using ImageNet examples.
    }
    \centering
    \begin{tabular}{|l|l|l|l|l|l|}
         \hline
         PGD & dataset & f1-score & precision & recall & acc \\
         \hline
         $\infty$-norm & test & 0.92 & 0.99 & 0.85 & 0.93 \\
         & train & 0.95 & 0.99 & 0.91 & 0.96 \\
         \hline
         $2$-norm & test & 0.96 & 0.98 & 0.94 & 0.96 \\
         & train & 0.99 & 0.97 & 0.99 & 0.99 \\
         \hline
    \end{tabular}
    \label{tab:mlp_metric}
\end{table}

\section{Conclusion} \label{sec:conclusion}

In this paper, we proposed a method that generates a one-dimensional input feature out of a raw image for being used as a rich, compact source of information for the detection of adversarial examples and ongoing attacks.
Our approach relies on computing the first digit distribution of an image's pixel values.
The assumption is that adversarial images do not follow Benford's law (BL) as natural images do.

Our approach comprises applying the Kolmogorov-Smirnov statistic between the first digit distribution of an image's pixels and the fixed theoretical distribution from BL after applying a suitable transformation to the given image.
Specifically, we have shown that the leading digit distribution of adversarial images generated by FGSM and PGD attack methods differs significantly from the corresponding distribution observed in unaltered images: the former deviates more compared to the latter, regarding BL.
This deviation tends to become higher as the magnitude of the perturbation increases, as shown for the FGSM attack.

Besides, one can use the proposed Feature for Adversarial Detection (FAD) to anticipate a potential undergoing attack since we have observed that, in many cases, the deviation given by the KS statistic reaches the separation threshold before the image becomes adversarial, that is, adversarial detection is feasible even before the attack is finished.

Future works include devising a sophisticated adversarial image detector based on the output of the KS statistic test, which one can employ as a low-dimensional input feature in conjunction with other metrics instead of the whole high-dimensional image. 
We have shown results from a logistic regression-based detector with only one input feature. However, it is possible to divide an image such that multiple KS statistics tests are obtained from the same input image, providing a more refined, informative view of the deviation caused by the attack perturbation.

Preliminary tests on the repeated application of the gradient transformation for a given image (e.g., two times) improved the results while employing the KL divergence instead of the KS test. This should be investigated in upcoming research.
Finally, different adversarial attacks, mainly black-box algorithms or those perturbing only a few pixels, should be tackled in future work.

\section*{Acknowledgments}
This work has been partially supported by CAPES - The Brazilian Agency for Higher Education (Finance Code 001), project PrInt CAPES-UFSC “Automation 4.0”.

\bibliography{refs}
\bibliographystyle{ieeetr}

\end{document}